\documentclass[journal]{IEEEtran}
\usepackage{cite}
\usepackage{amsmath,amssymb,amsfonts}
\usepackage{hhline}
\usepackage{tabularray}
\usepackage{algorithmic}
\usepackage{graphicx}
\usepackage{optidef}
\usepackage{enumitem} 
\usepackage{float}
\usepackage{svg}  
\usepackage{float} 
\usepackage{placeins} 
\usepackage[ruled,vlined]{algorithm2e}
\usepackage{amsmath}

\usepackage{multirow}
\usepackage{textcomp}
\usepackage{xcolor}
\usepackage{subcaption}
\usepackage{bm}
\def\BibTeX{{\rm B\kern-.05em{\sc i\kern-.025em b}\kern-.08em
    T\kern-.1667em\lower.7ex\hbox{E}\kern-.125emX}}
\usepackage{ragged2e}
\usepackage{booktabs}  
\usepackage{tabularx}
\usepackage{array}         
\usepackage{longtable}     
\usepackage{caption}       
\usepackage{graphicx}

\usepackage{tabularx}
\usepackage[table]{xcolor}
\usepackage{colortbl} 
\usepackage{tabularray} 
\usepackage{hhline} 
\usepackage{booktabs}

\newcolumntype{C}[1]{>{\centering\arraybackslash}p{#1}}
\newcolumntype{Y}{>{\RaggedRight\arraybackslash}X}

\usepackage{xcolor}
\usepackage{tabularx}  
\usepackage{array}
\usepackage{makecell}
\usepackage{booktabs} 

\newcolumntype{C}[1]{>{\centering\arraybackslash}m{#1}}

\newcolumntype{Z}{>{\centering\arraybackslash\hspace{0pt}}X}

\newcolumntype{L}{Z} 
\newcolumntype{M}{>{\hsize=1.00\hsize}Z} 
\newcolumntype{A}{>{\hsize=1.00\hsize}Z} 
\newcolumntype{D}{>{\hsize=1.00\hsize}Z} 

\definecolor{RowBlue}{RGB}{234,242,251}    
\definecolor{RowPeach}{RGB}{248,238,232}   
\definecolor{RowLav}{RGB}{241,237,247}     
\definecolor{RowMint}{RGB}{237,244,238}    
\definecolor{RowPink}{RGB}{244,236,239}    
\newcolumntype{C}[1]{>{\centering\arraybackslash}m{#1}}
\newcolumntype{L}{>{\raggedright\arraybackslash}X} 
\newcolumntype{Z}{>{\centering\arraybackslash}X}
\definecolor{ColAdv}{RGB}{255, 250, 205}  
\definecolor{ColLim}{RGB}{224, 255, 255}  

\ifCLASSINFOpdf
\else
\fi
\begin{document}
%
\title{Agentic Wireless Communication for 6G: Intent-Aware and Continuously Evolving Physical-Layer Intelligence}
%
%
%
\author{Zhaoyang Li, Xingzhi Jin, Junyu Pan~\IEEEmembership{Graduate Student Member, IEEE}, Qianqian Yang,~\IEEEmembership{Member, IEEE} and Zhiguo Shi,~\IEEEmembership{Fellow, IEEE}
\IEEEcompsocitemizethanks{
\IEEEcompsocthanksitem Zhaoyang Li, Xingzhi Jin, Junyu Pan, Qianqian Yang and Zhiguo Shi are with the College of Information Science and Electronic Engineering, Zhejiang University, Hangzhou,
China. (e-mails: \{zhaoyangli, 22560222, junyupan, qianqianyang20, shizg\}@zju.edu.cn).

\IEEEcompsocthanksitem This work is partly supported by the National Key R\&D Program of China under Grant No. 2024YFE0200802, by the National Natural Science Foundation of China (NSFC) under Grant Nos. 62293481, 62571487, and 62201505, by the Zhejiang Provincial Natural Science Foundation of China under Grant No. LZ25F010001. (Corresponding author: Qianqian Yang.)
}
}

\maketitle

\begin{abstract}
 As sixth-generation (6G) wireless systems continue to evolve, the increasing complexity of system functions and the diversification of service demands are driving a paradigm shift in system design, configuration, and operation—from rule-based control toward intent-driven autonomous intelligence. In 6G scenarios, user requirements are no longer characterized by single performance metrics such as throughput or reliability; instead, they are expressed as multi-dimensional objectives encompassing latency sensitivity, energy preferences, computational constraints, and service-level requirements, etc., which may evolve over time with environmental dynamics and user–network interactions. Consequently, accurately understanding both the communication environment and user intent is critical to enabling autonomous and sustainably evolving 6G communications. Large language models (LLMs), with their rich knowledge representations, strong contextual understanding, and cross-modal reasoning capabilities, provide a promising foundation for intent-aware network agents. Compared with conventional rule-driven or centrally optimized approaches, LLM-based agents can potentially integrate heterogeneous information sources and translate natural-language intents into executable network control and configuration decisions, thereby enabling more flexible and adaptive autonomy. Centered on the closed-loop process of intent perception, autonomous decision making, and network execution. This paper systematically investigates the role of agentic artificial intelligence in 6G wireless communications physical-layer and its realization pathways. We review representative 6G physical-layer tasks and analyze their limitations in supporting intent awareness and autonomous operation, identify application scenarios where agentic AI can demonstrate clear advantages, and discuss key challenges and enabling technologies related to multimodal perception, cross-layer decision making, and sustainable optimization. Finally, we present a case study of an intent-driven communication-link decision agentic AI, termed Agentic Communications (AgenCom). AgenCom adaptively constructs communication links under diverse user preferences and channel conditions, offering practical insights into building user-centric, autonomous, and sustainably evolving 6G wireless systems.
\end{abstract}


\section{Introduction}
\IEEEPARstart{T}{he} rapid development of wireless network technology and artificial intelligence is driving the evolution of sixth-generation (6G) communication towards greater intelligence and convergence. The vision of 6G is not only to achieve ultra-high data rates and ultra-low latency, but also to build an intelligent network capable of autonomous operation and seamless integration across air, space, and sea \cite{ITUR23M2160}. The core challenge in realizing this vision lies in its highly dynamic and complex operating environment: the system needs to continuously sense and process massive, heterogeneous, and rapidly changing data streams from intelligent edge devices, IoT sensors, and distributed computing platforms \cite{Boutouchent25INTENSE}. In this environment, traditional rule-based or centralized optimization methods have become inapplicable due to their lack of flexibility and real-time responsiveness. 
This directly drives a fundamental shift in system operation paradigms from ``performance-centric" to ``intent-driven". In this model, user needs are abstracted into dynamic intents that integrate multiple dimensions such as latency sensitivity, computational constraints, and service priorities \cite{Leivadeas23IBN}. Therefore, the ability to sense and interpret complex environments and user intents, and to make autonomous decisions and actions, becomes the core of 6G intelligence. This highlights the inevitability of AI agents as a key technology for achieving network self-governance and self-adaptation. 

Existing AI-enabled optimization approaches, including task-specific learning models and centralized control frameworks, have demonstrated promising results in isolated tasks such as resource allocation, power control, and beam selection. However, these solutions typically rely on predefined objectives, handcrafted state representations, or rigid control loops, limiting their adaptability in intent-diverse and rapidly changing environments. Moreover, their fragmented design often overlooks cross-layer dependencies and lacks a unified mechanism to translate high-level user intent into coordinated, end-to-end network actions, which is increasingly necessary as 6G systems become more heterogeneous and multi-objective.

Recent advances in large language models (LLMs) provide a promising opportunity to bridge this gap\cite{Yang26LLMWComm}. With strong contextual understanding and reasoning capabilities, as well as the ability to process heterogeneous inputs, LLMs can effectively connect human intent expressed in natural language with machine-executable network decisions. More importantly, when integrated into agentic AI systems, LLMs can serve as the decision-making core that enables a closed-loop ``perceive--reason--act--feedback'' workflow: the agent interprets user intent and environmental observations, invokes external tools to evaluate candidate actions, and continuously refines its behavior based on feedback, assisted by a memory module. As a result, LLM-driven agents offer a compelling pathway toward intent-aware, executable, and continually evolving control for autonomous 6G networks.

In this article, we explore how intent-aware agentic AI can serve as a foundational enabler for autonomous and sustainable 6G wireless systems. We begin by reviewing key 6G physical-layer tasks and examining current practices, along with their limitations in addressing intent diversity and growing system complexity. We then discuss the new opportunities unlocked by agentic AI and highlight representative application scenarios. Building on these insights, we propose an intent-aware agentic AI architecture, termed Agentic Communications (AgenCom), and analyze the key challenges and enabling technologies required for its practical deployment. Finally, we present a representative case study of AgenCom to demonstrate how an agentic system can dynamically adapt network behaviors to varying user intents and channel conditions, and we conclude by outlining future research directions toward fully autonomous, user-centric 6G wireless systems.

\section{Key Tasks and Design Challenges in 6G Wireless Networks}
In this section, we review several representative tasks in 6G networks and summarize mainstream research methodologies and typical solutions. We further highlight the pronounced limitations of existing approaches when confronted with multi-objective optimization, non-stationary environments, and intent-centric network requirements. The summary is shown in Table I and these observations lay the groundwork for introducing an agent-based paradigm in the subsequent sections.
\subsection{Communication Tasks in 6G Networks}
6G systems are expected to operate in highly dynamic propagation environments, while also facing substantial heterogeneity in user equipment and infrastructure capabilities, as well as increasingly diverse and evolving service demands. In this context, achieving intent-driven link autonomy fundamentally relies on intelligent physical-layer decision-making and adaptation mechanisms, including per-transmission modulation and coding selection, power control, beam management, and receiver configuration. Therefore, intent-driven autonomous communications require coherent and coordinated control within the physical layer to dynamically balance diverse performance objectives under complex channel conditions. With these considerations in mind, we next introduce several key physical-layer tasks in 6G communications.
\subsubsection{Beam Management}
Beam management aims to establish and continuously maintain high-quality directional links via beam discovery, alignment, tracking, and recovery, which is particularly critical for mmWave systems subject to mobility and blockage. Conventional solutions typically adopt codebook-based beam sweeping and hierarchical search for initial access, periodic probing for tracking, and threshold-triggered retraining for recovery; in multi-user settings, beam decisions are often coupled with scheduling via heuristic ranking based on RSRP/SINR. Deep-learning-based methods learn beam selection and tracking policies directly from measurements (e.g., CSI, location, and blockage indicators), reducing training overhead and improving robustness under non-stationary propagation; more recently, exploratory LLM-based studies have emerged to fuse multimodal context and provide high-level assistance for beam decision and prediction through stronger contextual modeling.

\subsubsection{Modulation and Coding Strategy Selection}
Modulation and coding scheme (MCS) selection jointly determines the modulation order and coding rate according to channel quality and service objectives, thereby controlling the reliability--throughput tradeoff. In practical systems, link adaptation is typically driven by CSI/CQI and implemented via predefined lookup tables (LUTs) over a discrete MCS set to meet a target BLER, together with standardized channel coding (e.g., LDPC and polar codes) and rule-based mechanisms such as OLLA and HARQ. Deep-learning-based approaches extract features from CSI and richer context (e.g., interference characteristics and ACK/NACK history) and learn complex nonlinear mappings to output more robust MCS decisions under specific objectives and channel conditions.

\subsubsection{Power Control}
Power control adjusts transmit power to balance link reliability, interference management, and energy consumption, which is essential for dense deployments and energy-constrained devices. Conventional methods include target-SINR control loops, interference-aware heuristics, and optimization under simplified interference/channel models, typically with fixed objectives and manually tuned parameters. Recently, multi-cell power control has been formulated as a reinforcement learning problem, where policies are learned from multi-cell observations and performance feedback to better handle non-stationary interference and improve decision flexibility.

\subsubsection{Adaptive Receiver Restoration}
Adaptive receiver restoration dynamically configures receiver-side processing---including channel estimation, equalization, detection, and decoding---to maintain robustness under time-varying channels, interference, and hardware impairments. Traditional receivers are largely model-based and modular, relying on fixed pipelines (e.g., MMSE estimation, linear/nonlinear equalization, and iterative decoding) with manually selected thresholds, update rates, and algorithmic settings, which can degrade under model mismatch and fast dynamics. Deep-learning-based receivers replace or augment key modules (e.g., learned estimators/detectors or unfolded networks) and can leverage closed-loop indicators (e.g., EVM and LLR statistics) for online adaptation, improving robustness in challenging conditions.

\subsubsection{Joint Physical-Layer Optimization}
Joint physical-layer optimization co-designs multiple coupled physical-layer decisions---such as beam/precoding, MCS, transmit power, pilot overhead, and receiver configuration---to achieve end-to-end objectives under dynamic channels and intent preferences. Conventional approaches often optimize these components in isolation or solve nonconvex problems under strong modeling assumptions, which can lead to locally optimal solutions and limited adaptability in the presence of practical impairments and non-stationary interference. Deep-learning-based methods attempt to formulate joint control as a reinforcement learning problem to exploit cross-module couplings; however, due to high-dimensional action spaces, complex constraints, and training stability issues, existing works typically optimize only a subset of modules or target a single objective. In contrast, LLM-based end-to-end planning and cross-module coordination for full physical-layer link construction remains relatively underexplored, and enabling executable structured decisions with stable closed-loop optimization is still an open challenge.

\subsubsection{Multi-task physical-layer Foundation Models}
Multi-task physical-layer foundation models aim to provide shared representations and reusable knowledge for multiple physical-layer tasks, enabling transfer across tasks, environments, and objectives. Traditional designs treat each physical-layer task as an independent pipeline with bespoke models and scenario-specific tuning, making cross-task reuse and rapid adaptation difficult; meanwhile, small-capacity deep learning models often struggle to share representations effectively across many tasks, limiting their ability to serve as truly general foundation models. Recently, large-model-based approaches have begun to leverage large-scale wireless channel datasets for self-supervised pretraining to learn more general channel representations, opening a new pathway toward transferable and scalable multi-task physical-layer foundation models.

\subsection{Limitations of Existing Approaches}Although conventional physical-layer solutions have achieved practical deployment, their inherent limitations become increasingly evident as 6G evolves toward intent-driven autonomy. In this section, we summarize the limitations of existing approaches from the perspectives of \emph{traditional model-driven designs}, \emph{learning-based methods}, and \emph{LLM-driven approaches}, while task- and method-specific comparisons are detailed in Table~\ref{tab1}.

\textit{Traditional methods:}
Most traditional physical-layer control mechanisms are built upon handcrafted rules, fixed codebooks or lookup tables, and offline-tuned thresholds or objective functions. While such designs are interpretable and engineering-friendly, they lack a unified mechanism to translate high-level user intent into coordinated and executable cross-module actions. Consequently, system objectives are often reduced to fixed performance metrics or static constraints, limiting the ability to dynamically adapt behavior in a user-centric and context-aware manner. Moreover, traditional procedures typically rely on simplified models, periodic updates, and strong operational assumptions, leading to slow responses to transient events and vulnerability under fast dynamics and non-stationarity (e.g., mobility and blockage, bursty interference, hardware impairments, and heterogeneous deployment conditions).

\textit{Deep learning based methods:}
Learning-based physical-layer methods can improve robustness against non-linear impairments and partially enhance adaptability. However, they often suffer from strong dependence on data availability and representativeness, and may experience significant performance degradation under distribution shifts and sim-to-real gaps. Additionally, the inherent "black-box" nature of neural networks leads to poor interpretability and auditability, making it difficult to verify performance and ensure safety in carrier-grade real-time pipelines.

\textit{LLM-driven methods:}
LLM-driven approaches introduce a promising capability to reason over heterogeneous observations and interpret user intent. Nonetheless, their practical deployment in physical-layer pipelines faces substantial challenges. These include prohibitive computational and latency overhead, resource constraints at the wireless edge, and difficulties in guaranteeing real-time responsiveness. More critically, probabilistic hallucinations and limited verifiability raise reliability and safety concerns, while the gap between high-level reasoning outputs and physically realizable, standard-compliant actions can hinder executability. Furthermore, privacy and security risks may arise when processing sensitive data or relying on cloud-based inference.

These limitations motivate new paradigms that can integrate heterogeneous observations, reason over user intent, and autonomously generate physically realizable and coordinated communication strategies.

\begin{table*}[!t]
\centering
\caption{Comparative landscape of traditional, learning-based, and LLM-driven approaches for physical-layer tasks.}
\label{tab1}
\scriptsize
\setlength{\tabcolsep}{3pt}
\renewcommand{\arraystretch}{1.5} 

\begin{tabularx}{\textwidth}{C{0.12\textwidth}|L|L|L|L}
\hline
\rowcolor{gray!15}
\textbf{Task Category} & \textbf{Description} & \textbf{Methods} & \textbf{Advantages} & \textbf{Limitations} \\
\hline

\rowcolor{RowBlue}
 & & \textbf{Traditional:} codebook sweeping, hierarchical search, periodic probing, threshold-triggered retraining; heuristic RSRP/SINR ranking. \cite{7134756} & \cellcolor{ColAdv} & \cellcolor{ColLim} \\
\rowcolor{RowBlue}
 & & \textbf{Deep learning based:} learns beam selection/tracking from CSI, location, and blockage indicators to cut training overhead. \cite{11016093} & \cellcolor{ColAdv} & \cellcolor{ColLim} \\
\rowcolor{RowBlue}
\multirow{-3}{=}[2em]{\centering\textbf{Beam Management}} &
\multirow{-3}{=}[2em]{Establishes and maintains directional links via beam discovery, alignment, tracking, and recovery under mobility/blockage.} & 
\textbf{LLMs:} fuses multimodal context and intent to assist beam prediction/probing strategy and scheduling-aware decisions. \cite{11308125} &
\cellcolor{ColAdv}\multirow{-6}{=}[-2em]{\raggedright
\textbf{Traditional Methods:} \par
$\bullet$ Ensure rigorous theoretical interpretability and performance guarantees. \par
$\bullet$ Deliver deterministic low-latency execution for seamless standardization. \par
$\bullet$ Require zero training overhead, enabling immediate deployment with stable reliability.\par \smallskip
} & 
\cellcolor{ColLim}\multirow{-6}{=}[-2em]{\raggedright
\textbf{Traditional Methods:} \par
$\bullet$ Exhibit poor robustness in non-stationary or complex physical dynamics. \par
$\bullet$ Constrained by rigid, single-objective decision-making architectures. \par 
$\bullet$ Suffer from high computational scaling as system dimensions increase.\par
\smallskip
} \\

\hhline{-|-|-|>{\arrayrulecolor{ColAdv}}-|>{\arrayrulecolor{ColLim}}-}
\arrayrulecolor{black}

\rowcolor{RowPeach}
 & & \textbf{Traditional:} CQI/CSI-driven LUT with BLER target; OLLA + HARQ; standardized coding (LDPC/polar). \cite{Shanthi2019} & \cellcolor{ColAdv} & \cellcolor{ColLim} \\
\rowcolor{RowPeach}
\multirow{-2}{=}[1em]{\centering\textbf{MCS Selection}} &
\multirow{-2}{=}[1em]{Selects modulation order and coding rate to trade off reliability and throughput given channel and service objectives.} & 
\textbf{Deep learning based:} predicts BLER or directly outputs MCS using CSI + interference + ACK/NACK history. \cite{8703432} &
\cellcolor{ColAdv} & 
\cellcolor{ColLim} \\
 
\hhline{-|-|-|>{\arrayrulecolor{ColAdv}}-|>{\arrayrulecolor{ColLim}}-}
\arrayrulecolor{black}

\rowcolor{RowMint}
 & & \textbf{Traditional:} target-SINR loops, interference-aware heuristics, simplified optimization with hand-tuned parameters. \cite{5956443}  & \cellcolor{ColAdv} & \cellcolor{ColLim} \\
\rowcolor{RowMint}
\multirow{-2}{=}{\centering\textbf{Power Control}} &
\multirow{-2}{=}{Adjusts transmit power to balance reliability, interference, and energy in dense and energy-constrained deployments.} & 
\textbf{Deep learning based:} learns multi-cell policies from observations + performance feedback to adapt to time-varying interference. \cite{11028592} &
\cellcolor{ColAdv}\multirow{-6}{=}[-1em]{\raggedright
\textbf{Deep Learning based Methods:} \par
$\bullet$ Provide strong robustness against non-linear impairments and interference. \par
$\bullet$ Enable autonomous adaptability within high-dimensional state spaces (e.g., massive MIMO). \par 
$\bullet$ Optimize signaling efficiency via compact latent channel representations. \par\smallskip
} & 
\cellcolor{ColLim}\multirow{-6}{=}[-1em]{\raggedright
\textbf{Deep Learning based Methods:} \par
$\bullet$ Experience significant performance degradation under distribution shifts. \par
$\bullet$ Lack transparency and auditability due to inherent black-box nature. \par 
$\bullet$ Depend heavily on scarce, verified, and high-quality real-world datasets. \par\smallskip
} \\

\hhline{-|-|-|>{\arrayrulecolor{ColAdv}}-|>{\arrayrulecolor{ColLim}}-}
\arrayrulecolor{black}

\rowcolor{RowLav}
 & & \textbf{Traditional:} modular pipelines (MMSE, linear/nonlinear EQ, iterative decoding) with fixed thresholds/rates. \cite{6334502} & \cellcolor{ColAdv} & \cellcolor{ColLim} \\
\rowcolor{RowLav}
\multirow{-2}{=}{\centering\textbf{Adaptive Receiver}} &
\multirow{-2}{=}[1em]{Dynamically configures estimation, equalization and decoding to maintain robustness under time-varying channels and interference.} & 
\textbf{Deep learning based:} learned estimators/detectors or unfolded networks; optional online adaptation using EVM/LLR. \cite{9159626} &
\cellcolor{ColAdv} & 
\cellcolor{ColLim} \\

\hhline{-|-|-|>{\arrayrulecolor{ColAdv}}-|>{\arrayrulecolor{ColLim}}-}
\arrayrulecolor{black}

\rowcolor{RowBlue}
 & & & \cellcolor{ColAdv} & \cellcolor{ColLim} \\
\rowcolor{RowBlue}
\multirow{-2}{=}{\centering \textbf{Joint Physical-layer Optimization}} &
\multirow{-2}{=}{Co-designs multiple coupled physical-layer decisions to achieve end-to-end performance objectives under dynamic channels.} &
\multirow{-2}{=}{\textbf{Deep learning based:} coordinated policies for subsets of modules due to scalability constraints. \cite{11345576}} &
\cellcolor{ColAdv} & 
\cellcolor{ColLim}\\

\hhline{-|-|-|>{\arrayrulecolor{ColAdv}}-|>{\arrayrulecolor{ColLim}}-}
\arrayrulecolor{black}

\rowcolor{RowPeach}
 & & & \cellcolor{ColAdv} & \cellcolor{ColLim} \\
\rowcolor{RowPeach}
 & & & \cellcolor{ColAdv} & \cellcolor{ColLim} \\
\rowcolor{RowPeach}
\multirow{-3}{=}{\centering \textbf{Multi-task Physical-layer Foundation Models}} &
\multirow{-3}{=}{Provides shared representations and reusable knowledge to support transfer across tasks, environments, and objectives.} & 
\multirow{-3}{=}{\textbf{LLMs based:} large-scale self-/weakly-supervised learning on wireless traces/sim data for shared representations. \cite{11036155}} &
\cellcolor{ColAdv}\multirow{-6}{=}[3em]{\raggedright
\textbf{LLMs based Methods:} \par
$\bullet$ Facilitate holistic multimodal context fusion and cross-layer orchestration. \par
$\bullet$ Transform abstract user intents into actionable network strategies. \par
$\bullet$ Support zero-shot task generalization through in-context learning capabilities.
} & 
\cellcolor{ColLim}\multirow{-6}{=}[3em]{\raggedright
\textbf{LLMs based Methods:} \par
$\bullet$ Impose prohibitive computational costs and inference latency. \par
$\bullet$ Trigger reliability concerns due to probabilistic hallucinations and safety risks. \par
$\bullet$ Expose privacy vulnerabilities during cloud-based processing of sensitive data.
} \\

\hline
\end{tabularx}
\end{table*}

\section{Agentic AI for Wireless Communications}
Unlike traditional learning methods, which are often offline-designed, task-specific, and goal-static, agentic AI emphasizes autonomy, adaptability, and interactivity, focusing on a closed-loop ``perception-decision-action” process. This enables agents to continuously interact with dynamic environments, integrate real-time feedback, and optimize strategies under multi-objective constraints, while consistently aligning with high-level intents. Based on the above characteristics, this section first introduces the general principles of intelligent agents, then explores the potential capabilities of agentic AI in wireless communications, and finally highlights its pivotal role in the evolution toward 6G.
\begin{figure*}[htbp]
    \centering
    \includegraphics[width=1\textwidth]{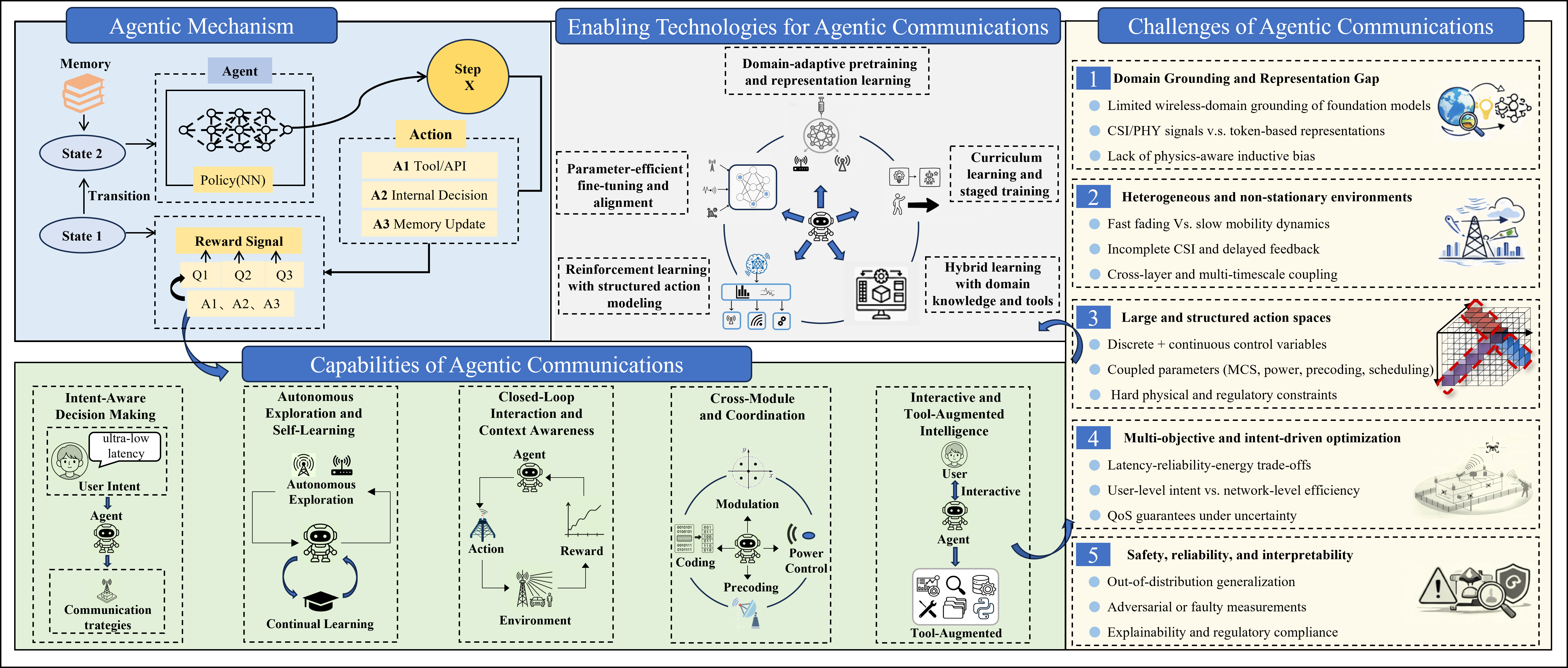} 
    \caption{Applications, mechanisms, and enabling techniques for agent.}
    \label{fig3}
\end{figure*}
\subsection{Mechanism of Agentic AI}
The core of agentic AI lies in building autonomous systems capable of perceiving the environment, making autonomous decisions, and executing actions to achieve goals. Its general working paradigm follows a classic ``perception-decision-action" closed loop, which can be enhanced through interaction with tools and environmental feedback, as shown in Fig.~\ref{fig3}. First, the agent acquires multimodal information from the environment through the perception module and constructs an internal state representation. Then, in the decision-making core, the agent generates a sequence of subsequent actions based on the current state, historical memory, and predetermined goals through planning or reasoning. This process can utilize external tools to enhance its capabilities. Next, the agent executes the selected action, thereby impacting the environment and triggering state changes. Feedback from the environment  is received by the agent and used to update its internal models or policies, enabling continuous learning and adaptation.

\subsection{Agentic AI for wireless
communication}
As shown in Fig.~\ref{fig3}, this subsection provides an overview of several promising research directions for intelligent agents in wireless communications.
\subsubsection{Intent-Aware Decision Making}

A defining characteristic of agentic AI is its ability to interpret and act upon high-level intents. In wireless communications, user requirements are increasingly expressed in terms of preferences and priorities---such as latency sensitivity, energy awareness, reliability levels, or computational constraints---rather than isolated performance metrics. Agentic AI enables the translation of such intent-oriented inputs into coordinated communication strategies, allowing network decisions to be driven by what users want rather than only by how the network operates. This intent-aware decision making is essential for tasks ranging from physical-layer link configuration to network slicing and service orchestration.

\subsubsection{Autonomous Exploration and Self-Learning}

Many communication tasks in 6G involve large and combinatorial decision spaces, as well as non-stationary environments caused by mobility, blockage, traffic fluctuations, and evolving network states. Agentic AI naturally supports autonomous exploration and continual learning through interaction with the environment, enabling agents to discover effective strategies beyond handcrafted rules or static optimization. Such self-learning capabilities are particularly beneficial for tasks like beam management, joint link adaptation, and energy-aware control, where optimal decisions depend on both instantaneous conditions and long-term trends.

\subsubsection{Closed-Loop Interaction and Context Awareness}

Agentic AI operates through closed-loop interaction, continuously observing system states, executing actions, and receiving feedback. This interaction-centric design allows agents to adapt their behavior based on rich contextual information, including channel dynamics, traffic patterns, device states, and historical outcomes. By maintaining internal state representations and leveraging temporal correlations, agentic AI can better handle uncertainty and delays, making it well suited for tasks that require sequential decision making, such as beam tracking, resource scheduling, and fault recovery.

\subsubsection{Cross-Module and Cross-Layer Coordination}

Traditional communication systems often rely on modular and layer-wise designs, where decisions are made independently within each functional block. Agentic AI provides a unifying framework to coordinate decisions across modules and layers by reasoning over coupled actions and shared objectives. This capability enables coherent strategy generation across physical-layer parameters (e.g., beam, modulation, power), RAN scheduling, edge computing decisions, and end-to-end orchestration. As a result, agentic AI can move beyond locally optimal solutions toward globally consistent and user-centric network behavior.

\subsubsection{Interactive and Tool-Augmented Intelligence}

Beyond passive optimization, agentic AI supports interactive intelligence by engaging with external tools, models, and knowledge sources. In wireless networks, such tools may include channel predictors, simulators, optimization solvers, or digital twins. By selectively invoking these tools and incorporating their outputs into the decision process, agents can enhance robustness and scalability while maintaining autonomy. This interaction between learning, reasoning, and tool usage is particularly relevant for complex tasks such as network planning, self-healing operations, and large-scale orchestration.

Together, these capabilities position agentic AI as a powerful enabler for intent-driven, autonomous, and sustainable 6G networks. By shifting the focus from task-specific optimization to goal-oriented and interactive intelligence, agentic AI offers a coherent framework to address a wide range of communication challenges discussed in the previous section and to support the advanced use cases envisioned for future wireless systems.
\section{Challenges and Enabling Technologies for Agentic AI in Wireless Systems}
Despite its strong potential, deploying agentic AI in wireless communication systems faces a number of fundamental challenges. These challenges stem not only from the complexity of wireless environments, but also from the mismatch between general-purpose AI models and domain-specific communication knowledge. As shown in Fig.~\ref{fig3}(c), this section first discusses key obstacles that limit the direct adoption of agentic AI in wireless networks, and then reviews enabling training and agent technologies that can facilitate scalable and domain-adaptive deployment.

\subsection{Key Challenges in Applying Agentic AI to Wireless Communications}

\subsubsection{Knowledge-domain mismatch}  
A central challenge lies in the mismatch between the knowledge embedded in general-purpose AI models and the highly specialized nature of wireless communication systems. Most large language models and foundation agents are trained on natural language and general-world data, whereas wireless networks are governed by domain-specific concepts such as channel models, signal processing pipelines, protocol constraints, and performance metrics. Bridging this gap requires agents to reason over physical-layer abstractions, mathematical models, and system-level constraints that are not explicitly represented in their pretraining data.

\subsubsection{Heterogeneous and non-stationary environments}  
Wireless environments are inherently dynamic, characterized by time-varying channels, mobility, interference, traffic fluctuations, and evolving network configurations. From an agent perspective, this leads to non-stationary state distributions and delayed or noisy feedback, which complicates learning stability and policy generalization, especially when agents are expected to operate across diverse scenarios.

\subsubsection{Large and structured action spaces}  
Many wireless decision problems involve combinatorial action spaces spanning multiple modules and layers, such as joint beam selection, modulation, power control, scheduling, and offloading. Naively treating such problems as flat action spaces results in poor scalability and slow convergence, posing a challenge for agent-based learning and exploration.

\subsubsection{Multi-objective and intent-driven optimization}  
Unlike single-metric optimization, wireless systems often need to balance conflicting objectives such as reliability, latency, throughput, energy efficiency, and computational cost. When user intent is incorporated, these objectives may further evolve over time. Designing agents that can robustly reason over such trade-offs and dynamically adjust priorities remains an open challenge.

\subsubsection{Safety, reliability, and interpretability}  
In operational wireless networks, decisions directly impact service quality and system stability. Agentic AI must therefore meet strict requirements on reliability, safety, and explainability. However, learning-based agents may exhibit unpredictable behavior during exploration or under distribution shifts, raising concerns about deployment in mission-critical scenarios.

\subsection{Agentic AI Technologies for Domain-Adaptive Deployment}

As shown in Fig.~\ref{fig3}, to address the aforementioned challenges, various training strategies and agent-centric techniques have been explored to adapt agentic artificial intelligence to wireless communication systems and enable scalable deployment. In this subsection, we provide a detailed discussion of these enabling technologies.

\subsubsection{Domain-adaptive pretraining and representation learning}  
Incorporating domain-specific data, such as channel realizations, protocol logs, and simulation traces, into pretraining can significantly reduce the knowledge gap between general-purpose models and wireless systems. Domain-adaptive representation learning allows agents to capture structural properties of wireless environments before task-specific optimization.

\subsubsection{Parameter-efficient fine-tuning and alignment}  
Parameter-efficient fine-tuning techniques, including low-rank adaptation and adapter-based methods, enable agents to acquire wireless-domain expertise without retraining large models from scratch. Such approaches are particularly suitable for adapting general agents to multiple wireless tasks while controlling computational overhead.

\subsubsection{Reinforcement learning with structured action modeling}  
To cope with large and structured action spaces, hierarchical decision making, factorized action representations, and sequence-based modeling can be employed. These techniques allow agents to decompose complex decisions into manageable sub-actions, improving learning efficiency and scalability.

\subsubsection{Hybrid learning with domain knowledge and tools}  
Agentic AI can be augmented with domain knowledge, analytical models, and external tools such as simulators, optimization solvers, and digital twins. By combining learning-based reasoning with model-based components, agents can achieve better robustness and sample efficiency while respecting system constraints.

\subsubsection{Curriculum learning and staged training}  
Training agents through curriculum learning or multi-stage pipelines---for example, warm-starting with heuristic policies or imitation learning before reinforcement learning---can stabilize convergence and improve performance in complex wireless environments. Such staged training strategies are particularly effective for intent-driven and multi-objective tasks.

Together, these enabling techniques provide a practical pathway for extending agentic AI across heterogeneous wireless tasks and layers. By systematically addressing domain mismatch, scalability, and reliability, they lay the foundation for deploying agentic AI as a core enabler of intent-driven, autonomous 6G networks.
\begin{figure*}[htbp]
    \centering
    \includegraphics[width=1\textwidth]{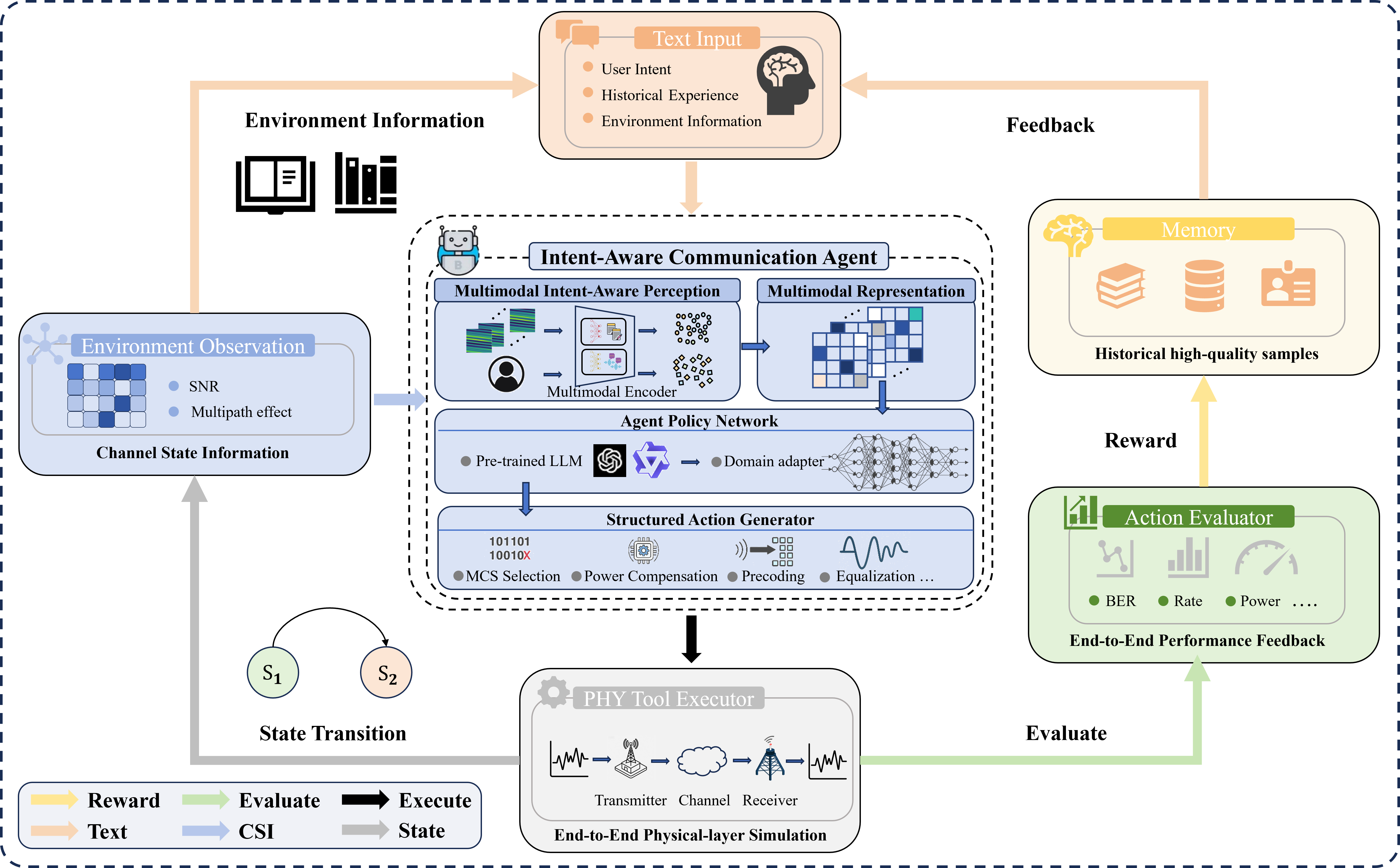} 
    \caption{Proposed intent-aware communication agent.}
    \label{fig2}
\end{figure*}

\section{Intent-Aware Agentic AI Architecture for Autonomous and Sustainable 6G Networks}
As show in Fig.~\ref{fig2}, to demonstrate the application of agentic AI in wireless communications, this section introduces an intent-aware communication agent, termed AgenCom, for adaptive physical-layer link construction. AgenCom jointly perceives CSI and user instructions expressed in natural language, and generates link configurations that align with diverse user intents. Notably, unlike conventional approaches that optimize individual physical-layer modules or a limited subset of modules in isolation, AgenCom treats the end-to-end link as the primary decision object and outputs a coherent and coordinated set of policies across all functional modules under the current channel conditions and user requirements. The overall architecture of the proposed AgenCom is detailed in the following section.

\subsection{Intent-Aware Perception and Representation through Multimodal Inputs}

AgenCom operates by jointly observing CSI and user intent expressed in natural language. CSI provides a concise and quantitative description of the physical-layer environment, while textual intent conveys high-level user preferences, such as reliability, throughput, or energy awareness, which are often difficult to capture through numerical indicators alone. By aligning these heterogeneous inputs within a shared representation space, AgenCom forms a unified perception of both environmental conditions and desired communication behavior. This multimodal perception allows AgenCom to associate physical-layer states with semantic intent, enabling it to reason not only about how the channel evolves, but also about why certain link behaviors are preferred under the current context. As a result, link decisions can be guided by user-centric objectives rather than fixed performance targets, providing a flexible interface for intent-driven communication.

\subsection{LLM-Based Domain-Adaptive Policy Network}

Built upon the fused multimodal representation, AgenCom employs a policy network to map multimodal context to executable link strategies. As shown in Fig.~\ref{fig2}, the policy network is centered around LLMs, which provide general-purpose reasoning over user intent and contextual information, and is augmented with a lightweight domain adapter to inject physical-layer priors, thereby improving controllability and executability under wireless constraints. With this LLM+adapter design, AgenCom can interpret natural-language intent instructions, reason over the feasible operating space, and generate decisions that satisfy system constraints.

\subsection{Structured Action Generation}

Based on the fused perception, AgenCom generates a complete link construction strategy by selecting a set of physical-layer configurations. Unlike conventional designs that optimize each module independently using predefined rules or lookup tables, AgenCom treats link construction as a unified decision problem and produces a coherent strategy that reflects dependencies across link components. To maintain scalability in the presence of combinatorial decision spaces, link decisions are structured as a sequence of interrelated sub-actions, allowing the agent to progressively refine the link configuration. This structured decision process enables AgenCom to capture causal relationships among physical-layer parameters and to adapt its behavior according to different user intents. Consequently, the same channel condition may lead to distinct link strategies depending on whether the user prioritizes reliability, throughput, or energy efficiency, highlighting the agent’s capability for personalized and context-aware link adaptation.

\subsection{Learning-Based Adaptation and Interaction}

As illustrated in Fig.~\ref{fig2}, AgenCom closes the loop by invoking a physical-layer tool executor to evaluate candidate link configurations under the current environment. The returned end-to-end performance metrics are used as reward signals for learning and policy refinement. Such tool-executed interaction provides a reliable, physics-grounded evaluation channel, which helps reduce the risk of producing infeasible or unsafe physical-layer actions and ensures the executability of the generated strategies. Moreover, AgenCom maintains a memory module that stores historical high-quality samples. By retrieving and reusing relevant experiences under similar channel and intent conditions, the memory module improves data efficiency and adaptation speed, while enhancing the consistency and stability of decision-making.

\section{Case Study}
To demonstrate the practical behavior of the proposed AgenCom, we build an end-to-end physical-layer communication environment in Sionna and conduct a case study on adaptive link construction using GPT-2 Medium (355M parameters). Sionna provides a differentiable and modular physical-layer pipeline, enabling systematic end-to-end evaluation of link design choices under realistic channel models and receiver processing. In our setup, each transmission instance corresponds to a complete physical-layer chain, including key modules such as channel coding, modulation, precoding, power control, channel estimation, and equalization. AgenCom jointly observes the current CSI and user intent specified in natural language, and accordingly outputs a physically feasible and coherent set of strategies to construct the transmission link, without relying on handcrafted rules or fixed lookup tables.

\subsection{Experimental Setup and Evaluation Protocol}

\subsubsection{User intent classes} We consider three representative intent classes that reflect common 6G service preferences: high-throughput, high-reliability, and energy-aware operation. These intents correspond to different priorities among throughput, error performance, and transmission power, and they serve as high-level control signals that guide the agent's link configuration.

\subsubsection{Action space} For each transmission instance, AgenCom selects a complete link strategy by choosing discrete configurations for key physical-layer modules, including channel coding schemes, coding rates, modulation orders, power compensation levels, precoding methods, and channel estimation and equalization techniques. The resulting strategy is then used to instantiate a full Sionna-based transmission chain, from which end-to-end performance feedback is generated to evaluate the effectiveness of the selected link configuration.
\begin{figure*}[htbp]
    \centering
    \includegraphics[width=1\textwidth]{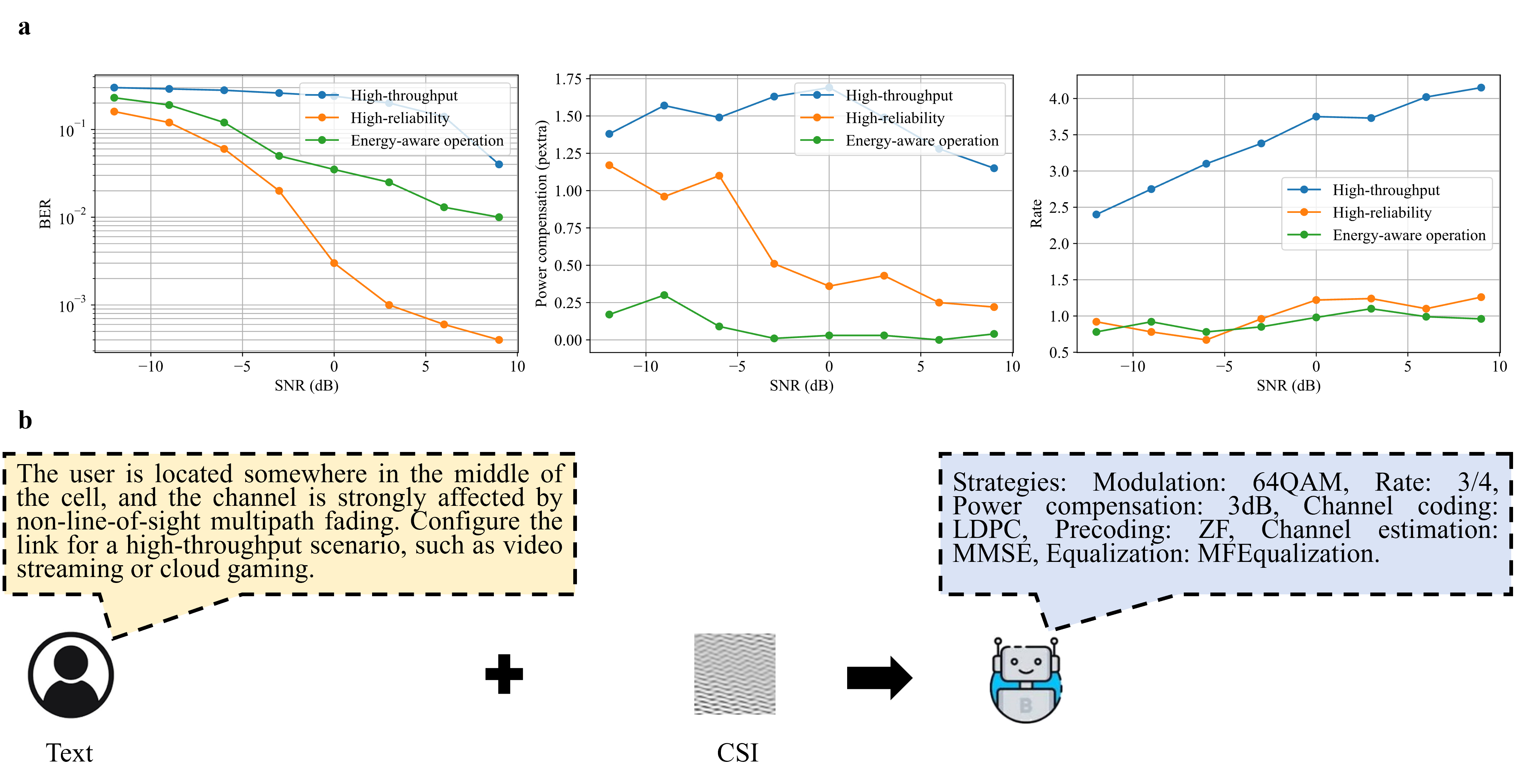} 
    \caption{Case studies of proposed Agent. (a) Comparison of output strategies for different user needs; (b) User Interaction Case.}
    \label{fig1}
\end{figure*}
\subsubsection{Data Generation}
To evaluate the agent’s adaptability under diverse propagation conditions, we generate CSI data using Sionna-RT with multiple parameter configurations, covering heterogeneous channel states and geometric scenarios. For each CSI sample, we further assign a user intent label in natural language to construct diverse settings where channel conditions and user preferences vary jointly, thereby testing the agent’s decision flexibility across heterogeneous environments and intent contexts.

\subsubsection{Performance Metrics}
We report three key metrics that jointly characterize user-centric trade-offs: (i) BER to quantify link reliability; (ii) achievable rate to represent throughput, defined as the ratio between the effectively delivered information over the link and the original data payload; and (iii) power compensation (denoted as $p_{\text{extra}}$) to measure the additional transmit-power overhead introduced by the selected strategy, where $p_{\text{extra}}$ is obtained by mapping the power-compensation setting to transmit power and then taking the natural logarithm relative to a reference power. Together, these metrics provide a clear view of how user intent shapes link strategy selection and end-to-end link behavior.

\subsubsection{Evaluation workflow} The end-to-end evaluation follows a consistent closed-loop procedure: (1) acquiring the current CSI and the user intent prompt; (2) having the agent output a link strategy based on these observations; (3) instantiating the corresponding Sionna physical-layer communication chain and running transmission simulations; (4) computing performance metrics including bit error rate, achievable rate, and power consumption; and (5) aggregating results across multiple trials and SNR levels. This closed-loop process explicitly tests whether the agent can translate user intent into executable physical-layer configurations and adapt its decisions in response to dynamic channel conditions.

\subsection{Results and Interpretation}

Fig.~\ref{fig1}(a) summarizes the performance trends under three user intent classes, clearly illustrating the intent-awareness and autonomous decision-making capability of AgenCom. As observed from the BER curves, the reliability-oriented intent consistently achieves the lowest BER across all SNR levels. This indicates that, under this intent, AgenCom tends to select conservative and robust physical-layer configurations, effectively prioritizing transmission reliability. Such reliability improvements come at an expected cost: compared with the throughput-oriented strategy, the achievable rate is reduced and higher power compensation is typically required, reflecting AgenCom’s willingness to allocate additional resources to maintain robust links.

Under the high-throughput intent, AgenCom achieves the highest data rates at nearly all SNR points, suggesting a preference for more aggressive modulation and coding configurations to maximize spectral efficiency. Correspondingly, the BER is higher than that of the reliability-oriented strategy, reflecting a deliberate trade-off aligned with the user’s throughput preference. Power compensation levels are also generally higher, indicating that AgenCom may allocate additional transmit power to sustain high-rate transmissions. In contrast, the energy-aware intent leads to systematically reduced power compensation, particularly in the medium-to-high SNR regime where satisfactory performance can be supported without excessive power expenditure. In this case, the achievable rate typically remains at a moderate level—higher than that of the reliability-oriented strategy but lower than that of the throughput-oriented strategy—while maintaining BER within a reasonable range. These trends confirm that when energy efficiency is prioritized, AgenCom learns to reduce power consumption by adopting more conservative link configurations rather than blindly maximizing throughput. Fig.~\ref{fig1}(b) shows a specific example of user interaction, where AgenCom receives the user's intent and CSI, and outputs the corresponding communication link decision.

This case study highlights several key advantages of the agent-based approach. First, by simply modifying the user intent, AgenCom can produce distinct and consistent link behaviors under identical channel conditions, providing a practical mechanism for personalized physical-layer adaptation. Second, instead of optimizing a single metric, AgenCom implicitly captures the multi-objective trade-offs among reliability, throughput, and power consumption, which are difficult to realize using fixed-rule designs. Third, by outputting a complete strategy across multiple physical-layer modules, AgenCom enables coordinated cross-module decision making, mitigating the fragmentation inherent in modular optimization. Overall, these results provide concrete evidence that intent-aware agents can serve as effective controllers for user-centric and adaptive link construction in future 6G wireless networks.

\section{Conclusion}
6G networks are envisioned to evolve from performance-centric connectivity toward user-centric, intent-driven autonomy, where communication, computing, and management decisions must be coordinated under highly dynamic environments and diverse service requirements. In this article, we reviewed representative tasks spanning the physical layer and higher network layers, and discussed why existing solutions face growing limitations when confronted with non-stationary conditions and multi-objective, intent-aware demands. We then highlighted agentic AI as a unifying paradigm that enables closed-loop perception--decision--action, autonomous exploration and self-learning, and interactive intent alignment. By bridging heterogeneous observations with user intent and generating coordinated actions across modules and layers, agentic AI offers a promising pathway to scalable and adaptive 6G intelligence. To ground the discussion, we presented an intent-aware link construction case study, showing that an agent can differentiate user preference classes and produce distinct link strategies that trade off reliability, throughput, and power in a consistent manner. Looking ahead, realizing agentic AI at scale will require advances in domain alignment, robust and safe learning under distribution shifts, structured action modeling for combinatorial decision spaces, and tool-augmented agents that can leverage simulators and digital twins. Addressing these challenges will be essential to translate the potential of intent-aware agents into practical, reliable, and sustainable 6G network deployments.

%
\IEEEpeerreviewmaketitle

\bibliographystyle{IEEEtran}
\bibliography{references}

@techreport{ITUR23M2160,
  author      = {{ITU-R}},
  title       = {{IMT-2030}: Framework and overall objectives of the future development of {IMT} for 2030 and beyond},
  institution = {International Telecommunication Union, Radiocommunication Sector (ITU-R)},
  type        = {Recommendation},
  number      = {M.2160-0},
  year        = {2023},
  month       = nov,
  url         = {https://www.itu.int/rec/R-REC-M.2160-0-202311-I/en},
  note        = {Accessed: 2026-02-08}
}

@article{Leivadeas23IBN,
  author    = {Leivadeas, Aris and Falkner, Matthias},
  title     = {A Survey on Intent-Based Networking},
  journal   = {IEEE Commun. Surv. Tutor.},
  year      = {2022},
  volume    = {25},
  number    = {1},
  month     = oct,
  pages     = {625--655},
  publisher = {IEEE}
}

@article{Boutouchent25INTENSE,
  author    = {Boutouchent, Akram and others},
  title     = {{6G-INTENSE}: Intent-Driven Native Artificial Intelligence Architecture Supporting Network-Compute Abstraction and
               Sensing at the Deep Edge},
  journal   = {IEEE Veh. Technol. Mag.},
  year      = {2025},
  month     = mar,
  volume    = {20},
  number    = {1},
  pages     = {44--54},
  publisher = {IEEE}
}

@article{Yang26LLMWComm,
  author    = {Yang, Ning and Fan, Mingrui and Wang, Wentao and Zhang, Haijun},
  title     = {Decision-Making Large Language Model for Wireless Communication: A Comprehensive Survey on Key Techniques},
  journal   = {IEEE Commun. Surv. Tutor.},
  year      = {2025},
  publisher = {IEEE}
}

@ARTICLE{7134756,
  author={Choi, Jinho},
  title={Beam Selection in mm-Wave Multiuser MIMO Systems Using Compressive Sensing}, 
  journal={IEEE
Trans. Commun.},
  volume={63},
  number={8},
  pages={2936-2947},
  year={Aug. 2015}
}

@ARTICLE{11016093,
  author={Rasheed, Iftikhar and Mostafa, Hala},
  journal={IEEE Trans. Veh. Technol.}, 
  title={DeepBeam: A Multi-Agent Deep Reinforcement Learning Framework for Predictive mmWave Beam Management in Dynamic V2X Networks}, 
  year={Nov. 2025},
  volume={74},
  number={11},
  pages={17854-17864}
  
  }

@ARTICLE{11308125,
  author={Lei, Jiahao and Li, Xiang and Wu, Chenbo and Fu, Qiang and Liu, Jiajia and Kato, Nei},
  journal={IEEE J. Sel. Areas
Commun.}, 
  title={{LLM-MM}: End-to-End Robust Multimodal Beam Prediction for 6G V2X Networks via MoE-LoRA Adaptation}, 
  year={2025},
  volume={},
  number={},
  pages={1-1}

}

@INPROCEEDINGS{5956443,
  author={Douros, Vaggelis G. and Polyzos, George C. and Toumpis, Stavros},
  booktitle={Proc. Veh. Technol. Conf.(VTC Spring), Budapest, Hungary}, 
  title={Negotiation-Based Distributed Power Control in Wireless Networks with Autonomous Nodes}, 
  year={May. 2011},
  volume={},
  number={},
  pages={1-5}
}

@ARTICLE{11028592,
  author={Zhang, Xiaoqing and Le, Van An and Kaneko, Megumi and Lui, John C.S. and Ji, Yusheng},
  journal={IEEE Trans. Veh. Technol.}, 
  title={Multi-Agent Deep Reinforcement Learning-Based Uplink Power Control in Cell-Free Massive MIMO With Mobile Users}, 
  year={Nov. 2025},
  volume={74},
  number={11},
  pages={17796-17811}
  
}

@ARTICLE{6334502,
  author={Wang, Chin-Liang and Shen, Po-Chung and Lin, Ying-Chang and Huang, Jia-Hong},
  journal={IEEE Trans. Commun.}, 
  title={An Adaptive Receiver Design for OFDM Systems Using Conjugate Transmission}, 
  year={Feb. 2013},
  volume={61},
  number={2},
  pages={599-608}
}

@ARTICLE{9159626,
  author={Yi, Xuemei and Zhong, Caijun},
  journal={IEEE Commun. Lett.}, 
  title={Deep Learning for Joint Channel Estimation and Signal Detection in OFDM Systems}, 
  year={Dec. 2020},
  volume={24},
  number={12},
  pages={2780-2784}
}

@ARTICLE{11036155,
  author={Yang, Tingting and Zhang, Ping and Zheng, Mengfan and Shi, Yuxuan and Jing, Liwen and Huang, Jianbo and Li, Nan},
  journal={IEEE Netw.}, 
  title={WirelessGPT: A Generative Pre-Trained Multi-Task Learning Framework for Wireless Communication}, 
  year={Sep. 2025},
  volume={39},
  number={5},
  pages={58-65}
}

@ARTICLE{11345576,
  author={Zheng, Xufei and Xiao, Han and Jin, Shi and Wang, Zhiqin and Tian, Wenqiang and Liu, Wendong and Cao, Jianfei and Shen, Jia and Shi, Zhihua and Zhang, Zhi and Yang, Ning},
  journal={IEEE J. Sel. Areas Commun.}, 
  title={AI-Native 6G Physical Layer with Cross-Module Optimization and Cooperative Control Agents}, 
  year={2026},
  volume={},
  number={},
  pages={1-1}
}

@Article{Shanthi2019,
author={Shanthi, K. G.
and Manikandan, A.},
title={An Improved Adaptive Modulation and Coding for Cross Layer Design in Wireless Networks},
journal={Wireless Pers. Commun.},
year={Sep. 2019},
day={01},
volume={108},
number={2},
pages={1009-1020}
}

@ARTICLE{8703432,
  author={L. Zhang and J. Tan and Y.-C. Liang and G. Feng and D. Niyato},
  journal={IEEE Trans. Wireless Commun.}, 
  title={Deep Reinforcement Learning-Based Modulation and Coding Scheme Selection in Cognitive Heterogeneous Networks}, 
  year={Jun. 2019},
  volume={18},
  number={6},
  pages={3281-3294}
}

\end{document}